\let\origvec\vec
\documentclass[runningheads]{llncs}

\let\vec\origvec
\usepackage{graphicx}
\usepackage{comment}
\usepackage{amssymb}
\usepackage{amsmath}
\usepackage{algorithm}		
\usepackage{algpseudocode} 	
\usepackage{subfig}

\usepackage{enumerate}
\usepackage{stfloats}
\usepackage{lineno}
\usepackage[hidelinks]{hyperref} %

\usepackage{sidecap}
  
\usepackage[textsize=footnotesize,color=yellow,colorinlistoftodos]{todonotes}

\begin{document}
\title{Self-organized Collective Motion with a Simulated Real Robot Swarm\thanks{This work was supported by EPSRC Impact Acceleration Account (EP/R511626/1).}}
%
%
\author{Mohsen Raoufi\inst{1} \and
Ali Emre Turgut\inst{2} \and
Farshad Arvin\inst{3}}
\authorrunning{M. Raoufi et al.}
%
\institute{\inst{1} Department of Aerospace Engineering, Sharif University of Technology, Tehran, Iran, email: \email{mohsen\_raoufi@ae.sharif.edu}
\\
\inst{2} Mechanical Engineering Department, 
Middle East Technical University,
06800 Ankara, Turkey, email: \email{aturgut@metu.edu.tr}
\\
\inst{3} School of Electrical and Electronic Engineering, The University of Manchester, M13~9PL, Manchester, UK, email: \
\email{farshad.arvin@manchester.ac.uk}}
\maketitle              
\begin{abstract}
Collective motion is one of the most fascinating phenomena observed in the nature. In the last decade, it aroused so much attention in physics, control and robotics fields. In particular, many studies have been done in swarm robotics related to collective motion, also called flocking. In most of these studies, robots use orientation and proximity of their neighbors to achieve collective motion. In such an approach, one of the biggest problems is to measure orientation information using on-board sensors. In most of the studies, this information is either simulated or implemented using communication. In this paper, to the best of our knowledge, we implemented a fully autonomous coordinated motion without alignment using very simple Mona robots. We used an approach based on Active Elastic Sheet (AES) method. We modified the method and added the capability to enable the swarm to move toward a desired direction and rotate about an arbitrary point. The parameters of the modified method are optimized using TCACS optimization algorithm. We tested our approach in different settings using Matlab and Webots.
\keywords{Swarm Robotics \and Collective Motion \and Coordinated Motion \and Flocking \and Self-organized.}
\end{abstract}
\section{Introduction}
\label{sec:Intro}
Collective motion (CM) is an eye-catching demonstration of a more broad phenomenon called collective behaviour~\cite{vicsek2001fluctuations}. CM is observed in diverse domains such as: physics~\cite{kudrolli2008swarming}, chemistry~\cite{lewandowski2009anisotropic}, from micro-size creatures such as bacteria colonies~\cite{sokolov2007concentration}, cells~\cite{arboleda2010movement}, macro-molecules~\cite{schaller2010polar} to handful macro-size examples of fish~\cite{ward2008quorum}, birds~\cite{bajec2009organized,ballerini2008interaction}, even humans~\cite{moussaid2011simple} and to non-living systems~\cite{suematsu2010collective}. The principle feature of CM is that each individual agent behaves based on the interaction between its neighbors; resulting in polarized motion of the group. This interaction might be either simple (attraction/repulsion) or complex (combinations of simple interactions)~\cite{vicsek2012collective}.\par
%
%
%
%
%
In most of the works on CM, agents use both orientation and proximity information of their neighbors~\cite{gregoire2004onset,couzin2002collective}. In contrast to these works, some works introduced methods that do not rely on this information. An individual-based model is defined in~\cite{romanczuk2009collective}, where the interaction between neighbor agents is defined only by escape-pursuit behavior. In~\cite{grossman2008emergence}, inelastic collisions between isotropic agents, and in~\cite{menzel2012soft} a pairwise repulsive force between deform-able, self-propelled particles are introduced as interaction mechanism regardless of the orientation of neighbors. These methods leave an indirect effect on individual agents, and as a result, enable them to establish aligning interaction in an effective, implicit manner. By so doing, not only the swarm achieves CM, but also the reduced dynamics reaches consensus in the heading direction of agents. An example of this implicit aligning is~\cite{ferrante2013collective}, in which, Ferrante et al. identified a new elasticity-based approach for achieving collective motion and illustrated the behavior by introducing an Active Elastic Sheet (AES) model. In their proposed model, the motion of an individual is determined by only attraction-repulsion forces. This feature is very useful in practical cases, such as a swarm of robots in which a specific robot have not any information of its neighbors' orientation~\cite{arvin2016}.\par

%
%
%
In this paper, we improve the AES model presented in~\cite{ferrante2013collective}, and optimize some of the parameters of the proposed algorithm using a heuristic optimization algorithm. To illustrate the improved performance of the model, various simulations with different assumptions are performed. In addition, some simulations are conducted in Webots~\cite{michel2004cyberbotics} using the model of Mona robot~\cite{arvin2018mona}. Finally, We implement the AES model on a swarm of simulated real Mona robots for the first time. To the best of our knowledge, this is the first implementation of fully autonomous CM using such a simple robotic platform without alignment information.\par
This paper is organized as follows: in the Sect. \ref{sec:collective motion control}, we introduce the concept of AES mechanism to achieve CM, and the modifications are addressed there. Then, the optimization algorithm is discussed and the objective function is defined in Sect. \ref{sec:optimization}. We mention the settings of simulations in the Sect. \ref{sec:Exp. Setup}. In Sect. \ref{sec:Results}, the results are shown and we discuss them. Sect. \ref{sec:Conclusion} closes the paper with a conclusion about the current work.
%
%
%
%
\section{Collective Motion Control}
\label{sec:collective motion control}
In this section, the dynamic model of AES along with some modifications is presented. The model is derived based on a simple two-dimensional active elastic sheet mechanism~\cite{ferrante2013collective}. After defining the basic mechanism, the numerical dynamics of particles and modifications will be elaborated upon.
\subsection{Active Elastic Sheet Model and Numerical Dynamics}
\label{subsec:AESModel}
\begin{figure}[!t]
\centering
\includegraphics[width=0.25\textwidth]{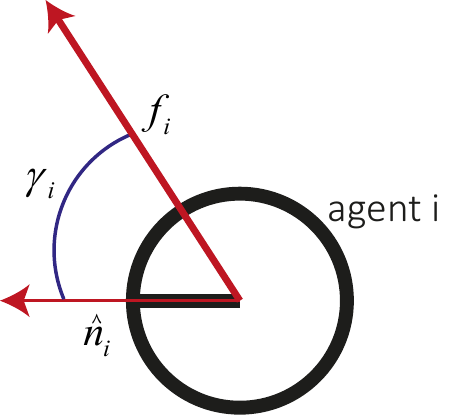}
\caption{The angle between force and heading vectors}
\label{fig: heading}
\end{figure}

Each agent $i$ out of $N$ agents, is affected by the attraction-repulsion force of its neighbors. This force will leave an effect on both the linear velocity $\boldsymbol{\dot{x}}_{i}$ and the rotational velocity $\dot{\theta}_{i}$ of the agent. This effect is formulated in continuous-time form as:
\begin{equation}
\label{eq: velocity}
\boldsymbol{\dot{x}}_{i}(t) = \left\{ v_{0}(t) + \alpha \left[ \left( \boldsymbol{f}_{i}(t) + D_{r}\boldsymbol{\hat{\xi}}_{r}(t) \right) \centerdot \boldsymbol{\hat{n}}_{i}(t)\right] \right\} \boldsymbol{\hat{n}}_{i}(t) 
\end{equation}
%
\begin{equation}
\label{eq: rot velocity}
{\dot{\theta}}_{i}(t) = \beta \left\{ \left[ \boldsymbol{f}_{i}(t) + {D_{r}}\boldsymbol{\hat{\xi}}_{r}(t) \right] \centerdot \boldsymbol{\hat{n}}_{i}^{\perp}(t)\right\} + {D_{\theta}}\boldsymbol{\xi}_{\theta}(t)
\end{equation}
where, $v_{0}$ is the forward biasing speed, and parameters $\alpha$ and $\beta$ are the inverse transitional and rotational coefficients, respectively. The unit vector $\boldsymbol{\hat{n}}_{i}(t)$ is the heading vector which is parallel to the heading direction of agent $i$, and $\boldsymbol{\hat{n}}_{i}^{\perp}(t)$ is perpendicular to it. 
\begin{equation}
\label{eq: heading}
\boldsymbol{\hat{n}}_{i}(t) = \begin{bmatrix} \cos(\theta_{i}(t))\\ \sin(\theta_{i}(t)) \end{bmatrix}
\end{equation}
The term $\boldsymbol{f}_{i}\centerdot \boldsymbol{\hat{n}}_{i}^{\perp}$ of Eq. (\ref{eq: rot velocity}) indicates that the angle between force and heading vectors ($\gamma_i$) defines the rotation speed of the agent (Fig.~\ref{fig: heading}). So, the rotation speed will be zero if two vectors $\boldsymbol{f}_{i}$ and $\boldsymbol{\hat{n}}_{i}$ are parallel (i.e. $\boldsymbol{f}_{i}$ and $\boldsymbol{\hat{n}}_{i}^{\perp}$ are perpendicular.) \par
Agents are subject to measurement and actuation noises ${D_{r}}\boldsymbol{\hat{\xi}}_{r}(t)$ and ${D_{\theta}}\boldsymbol{\xi}_{\theta}(t)$; in which, $\boldsymbol{\hat{\xi}}_{r}(t)$ is a randomly oriented unit vector and $\boldsymbol{\xi}_{\theta}(t)$ is a random variable with standard normal distribution. ${D_{r}}$ and ${D_{\theta}}$ are noise strength coefficients. Inducing total force vector $\boldsymbol{f}_{i}(t)$ over agent $i$ at time $t$ can be calculated by the summation of linear spring-like forces, which link agent $i$ to its interacting neighbor $j$, member of set $S_{i}$:
\begin{equation}
\label{eq: force}
\boldsymbol{f}_{i}(t) = \sum_{j \in S_{i}} {-\dfrac{k}{l_{ij}} \left( \lVert {\boldsymbol{r}_{ij}(t)} \rVert - l_{ij} \right) \dfrac{\boldsymbol{r}_{ij}(t)}{\lVert \boldsymbol{r}_{ij}(t) \rVert}}
\end{equation}
\begin{equation}
\label{eq: r_ij}
\boldsymbol{r}_{ij}(t) = \boldsymbol{{x}}_{j}(t) - \boldsymbol{{x}}_{i}(t)
\end{equation}
${l_{ij}}$ is equilibrium distance (or alternatively called natural length) of the spring that links agent $i$ and $j$. 
%
%
In order to simulate the model, we need to numerically integrate the dynamic equations of motion. Integrating Eq. (\ref{eq: velocity}) and (\ref{eq: rot velocity}) using Euler method at time step $t$, results in the governing discrete-time kinematic equations of each particle which is expressed by the following equations: 
%
%
\begin{equation}
\label{eq: linear 1}
\boldsymbol{x}_{i}^{t+1} = \boldsymbol{x}_{i}^{t} + \left\{ v_{0}\boldsymbol{\hat{n}}_{i}^{t} + \alpha \left[ \left( \boldsymbol{f}_{i}^{t} + \dfrac{D_{r}}{\sqrt{\Delta t}}\boldsymbol{\hat{\xi}}_{r}^{t} \right) \centerdot \boldsymbol{\hat{n}}_{i}^{t}\right] \boldsymbol{\hat{n}}_{i}^{t} \right\} \Delta t
\end{equation}
\begin{equation}
\label{eq: rotation 1}
{\theta}_{i}^{t+1} = {\theta}_{i}^{t} + \left\{ \beta \left[ \left( \boldsymbol{f}_{i}^{t} + \dfrac{D_{r}}{\sqrt{\Delta t}}\boldsymbol{\hat{\xi}}_{r}^{t} \right) \centerdot \boldsymbol{\hat{n}}_{i}^{\perp,t}\right] + \dfrac{D_{\theta}}{\sqrt{\Delta t}}\boldsymbol{\xi}_{\theta}^{t} \right\} \Delta t
\end{equation}
here, $\boldsymbol{x}_{i}^{t}$ and ${\theta}_{i}^{t}$ are the position and orientation of agent $i$ at step time $t$, respectively, and $\Delta t$ is the numerical time-step interval of integration. 
It is noteworthy that the velocity vector in Eq. (\ref{eq: velocity}) points parallel to the heading of the agent which means it is assumed that each agent can only turn, and move forward/backward. Modeling agents with 2 degrees of freedom provides many advantages to robotic applications in which the robots are not able to move omni-directionally, e.g. Mona~\cite{arvin2018mona} robot.\par
\subsection{Modifications to AES}
\label{subsec:modifications to AES}
In this section, we present some modifications to AES model, which result in collective linear and rotational motion of swarm. Firstly, we add a stimulating force to compel the swarm to move toward a specific direction. This modification leads to faster convergence of collective motion. Furthermore, the desired direction of swarm movement is also achievable by such modification. To model this auxiliary inducing force, at a certain time, we simply added the auxiliary force $\boldsymbol{f}_{aux,\ l}$ to the induced force of Eq. (\ref{eq: force}). This force is parallel to the desired velocity unit vector $\boldsymbol{\hat{v}}_d$ and its magnitude is defined by a weighting parameter $w_l$ as follows:
\begin{equation}
\label{eq: f_aux_l}
\boldsymbol{f}_{aux,\ l} = w_l \boldsymbol{\hat{v}}_d
\end{equation} \par
The second modification enables the swarm to rotate collectively by adding another auxiliary force. This force is also related to another desired linear velocity which itself is proportional to the rate of rotation. According to the kinematics of 2D rotation, the linear speed of each agent is proportional to its distance with respect to the center of rotation $\boldsymbol{x}_c$. Thus, pointed in the following equation, the linear velocity of rotation $\boldsymbol{v}_r$ is related to $\boldsymbol{r_{ic}}$, the position vector of agent $i$ wrt the center of rotation, and $\omega$ is the rate of rotation. $\boldsymbol{r_{ic,x}}$ and $\boldsymbol{r_{ic,y}}$ are the x- and y-component of $\boldsymbol{r_{ic}}$.
\begin{equation}
\label{eq: v_r}
\boldsymbol{v}_r = \omega \begin{bmatrix} \boldsymbol{r_{ic,y}}\\ -\boldsymbol{r_{ic,x}} \end{bmatrix}
\end{equation}
\begin{equation}
\label{eq: r_ic}
\boldsymbol{r_{ic}} = \boldsymbol{x_i} - \boldsymbol{x_{c}}
\end{equation}
Rotation center $\boldsymbol{x_{c}}$ is an arbitrary point, yet it is common to consider it on the center of swarm. Hence, same as the previous approach, we calculated the modified force for rotation by the following equation:
\begin{equation}
\label{eq: f_aux_r}
\boldsymbol{f}_{aux,\ r} = w_r \boldsymbol{v}_r
\end{equation}
where, $w_r$ is the weighting parameter of induced force for swarm rotation. It is worth mentioning that $\boldsymbol{f}_{aux,\ r}$ should be calculated for each agent, but $\boldsymbol{f}_{aux,\ l}$ is a universal force and is same for all agents. The linearity feature of the aforementioned phenomena enables us to add both terms together and form the total auxiliary force, which still satisfies both motions. So, the final modified inducing force is:
\begin{equation}
\label{eq: modified force}
\boldsymbol{f} = \boldsymbol{f}_{neighors} +  \boldsymbol{f}_{aux,\ l} + \boldsymbol{f}_{aux,\ r}
\end{equation}
$\boldsymbol{f}_{neighors}$ is exactly the same as Eq. (\ref{eq: force}), but we added the subscript to show that this is the force which is defined by the influence of interacting neighbors. 
\subsection{Degree of Alignment}
\label{subsec:DoA}
Since the symmetry is a principal feature for achieving flocking behavior, a criteria is introduced to measure it. This metric is defined by a specific system variable, called the degree of alignment. One of the most common, and widely used definition is represented in~\cite{vicsek2012collective}. In this case the variable $\psi$ is the average normalized velocity vectors:
\begin{equation}
\label{eq: DoA 1}
\psi = \dfrac{1}{N v_0} \left\lVert{\sum_{i=1}^{N} \boldsymbol{v}_{i}}\right\rVert 
\end{equation}
where $N$ is the total number of the agents, $v_0$ is the average absolute velocity, and $\boldsymbol{v}_i$ is the velocity vector of the agent $i$. However, in methods with 2-DoF agents, like AES, as previously mentioned, the velocity vector is parallel to the heading direction. So, the degree of alignment of the system is defined as~\cite{ferrante2013collective}:
\begin{equation}
\label{eq: DoA 2}
\psi = \dfrac{1}{N} \left\lVert{\sum_{i=1}^{N} \boldsymbol{\hat{n}}_{i}}\right\rVert 
\end{equation}
Because our purpose is to simulate the AES method on robots, we added another criterion which is defined from a control viewpoint. In this respect, the alignment of each particle $\psi_{i}$ is defined by its desired direction of movement, i.e. an agent is called aligned when its orientation is parallel to its desired direction of motion. Referred to Fig.~\ref{fig: heading}, the alignment is defined by the angle $\gamma_i$ as follows:
\begin{equation}
\label{eq: DoA i 3}
\psi_i = \left\lvert{cos(\gamma_i)}\right\rvert = \left\lvert{\dfrac{\boldsymbol{f}_i \centerdot \boldsymbol{\hat{n}}_{i}}{\left\lVert{\boldsymbol{f}_i}\right\rVert}}\right\rvert
\end{equation}
In this definition, $\psi_i$ is between zero (agent's orientation is perpendicular to the force, $\gamma_i=\pi/2$) and one (agent is directed toward the force, $\gamma_i = 0$ or $\gamma_i = \pi$). Note that, as the norm of the unit vector $\boldsymbol{\hat{n}}_{i}$ is equal to one, it is eliminated from the denominator. Therefore, the degree of system alignment is calculated by the following equation:
\begin{equation}
\label{eq: DoA 3}
\psi = \dfrac{1}{N} \left\lVert{\sum_{i=1}^{N} \psi_i}\right\rVert
\end{equation}
Mathematically, the value of this variable is between zero (asymmetric, disordered, random orientation) and one (fully symmetric orientation). 
%
%
%
%
%
%
\section{Optimization}
\label{sec:optimization}
In this section, the parameter optimization of the algorithm will be introduced. For so doing, we utilized a heuristic optimization algorithm called Tabu Continuous Ant Colony System (TCACS)~\cite{karimi2010continuous}. TCACS combines two algorithms: Continuous Ant Colony System (CACS)~\cite{pourtakdoust2004extension}, and Tabu Search (TS)~\cite{siarry1997fitting}. We will mention a brief description of the algorithm in the next subsection. Our aim is to find the optimal decision parameters for AES model with a specific setup configuration. The decision parameters to be tuned are $\alpha$, $\beta$, and $k$.
\subsection{TCACS Optimization Algorithm}
\label{subsec:TCACS}
\begin{figure}[!t]
\centering
\includegraphics[width=0.4\textwidth]{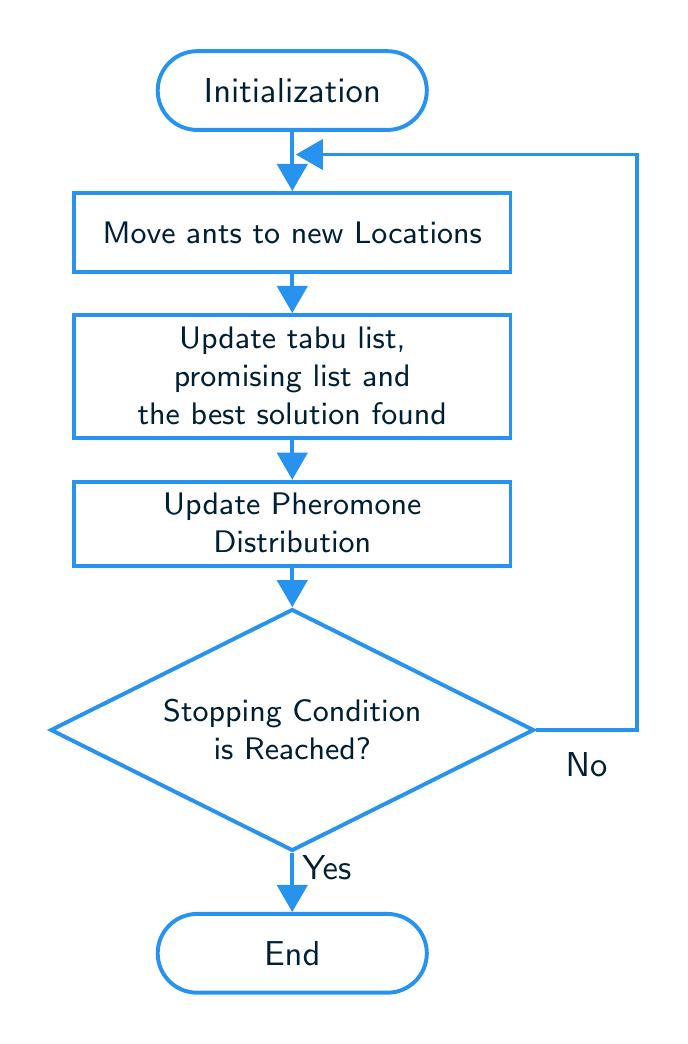}
\caption{General Flowchart of TCACS}
\label{fig: TCACS Flow Chart}
\end{figure}

The structure of the algorithm is very similar to CACS; however, the advantages of tabu concept are also added. Borrowing the concept of tabu balls (meta-spherical shapes in the decision-space), prevents the ants choosing their destination within the tabu regions.  
The flowchart of TCACS is illustrated in Fig.~\ref{fig: TCACS Flow Chart}. \par
In this paper, we considered the parameters of TCACS  as $N_{ants}=15$, $\gamma=0.5$, $m=2$ (which are number of ants, weighting factor, and PCA factor, respectively) and used Roulette as the weighting strategy. 
\subsection{Objective Function}
\label{subsec:objective function}
One of the most important steps in optimization is the definition of objective function, which can leave a considerable effect on the performance of the algorithm to solve the problem. 
Various objective functions can stand as candidates, depending on what the aim of the problem is and which conditions needs to be satisfied. In case of collective motion using the modified AES method, we have two main objectives:
\begin{itemize}
  \item To minimize the total induced force, which, generally speaking, means that agents are at their ideal position according to the elastic force of the sheet. Satisfying this objective will result in maintaining the shape of swarm same as its natural shape.
  \item To maximize the rate of convergence, which will be reached when the orientation of agents are about the same as the desired ones. To fulfill this objective, the optimization algorithm will try to maximize the degree of system alignment.
\end{itemize}
Although the defined problem is multi-objective, we combine both previous functions, multiplied by their weights, into a single objective function. This will help us to simplify and solve the parameter tuning problem in a more straight-forward manner. \par
It is notable that the system is subject to noise; consequently, the objective function gets stochastic, and it is necessary to reduce the effect of noise on it, otherwise, the performance of optimization will be reduced. Therefore, we calculate the average of multiple Monte Carlo simulations as the output of the objective function. With this in mind, the objective function is as follows:
\begin{equation}
\label{eq: obj F}
J = \dfrac{1}{N_{MC}}\sum_{m=1}^{N_{MC}}{ \left\lbrace \sum_{t=0}^{t_f}{  \left[  w_1 \left( \sum_{i=1}^{N}{\left\lVert{\boldsymbol{f}_i^t}\right\rVert} \right)  + w_2 {\psi^t}  \right] } \right\rbrace }
\end{equation}
In which, the first term of the RHS is calculated by Eq. (\ref{eq: modified force}), and the second term by Eq. (\ref{eq: DoA 3}). Additionally, $N_{MC}$ is the number of Monte Carlo simulations which is set to 10, and $t_f$ is the final time of simulation. Considering the fact that the first objective is a minimization problem, and the second objective is a maximization problem, we determined $w_1$ (the weighting parameter of the first function) positive and $w_2$ negative. As the result, the optimization can be alternatively called minimization, and the objective function is a \emph{cost} function. Determining the value of weighting parameters depends mainly on the purpose of the optimization, however, a common initial setting is normalizing the order of each term; so, we considered them as $w_1={1}/{N}$, and $w_2 = -1$.
%
%
%
%
\section{Experimental Setup}
\label{sec:Exp. Setup}



%
%
%
\begin{figure}[tb]
\centering
\subfloat[]{
\includegraphics[height=3.5cm]{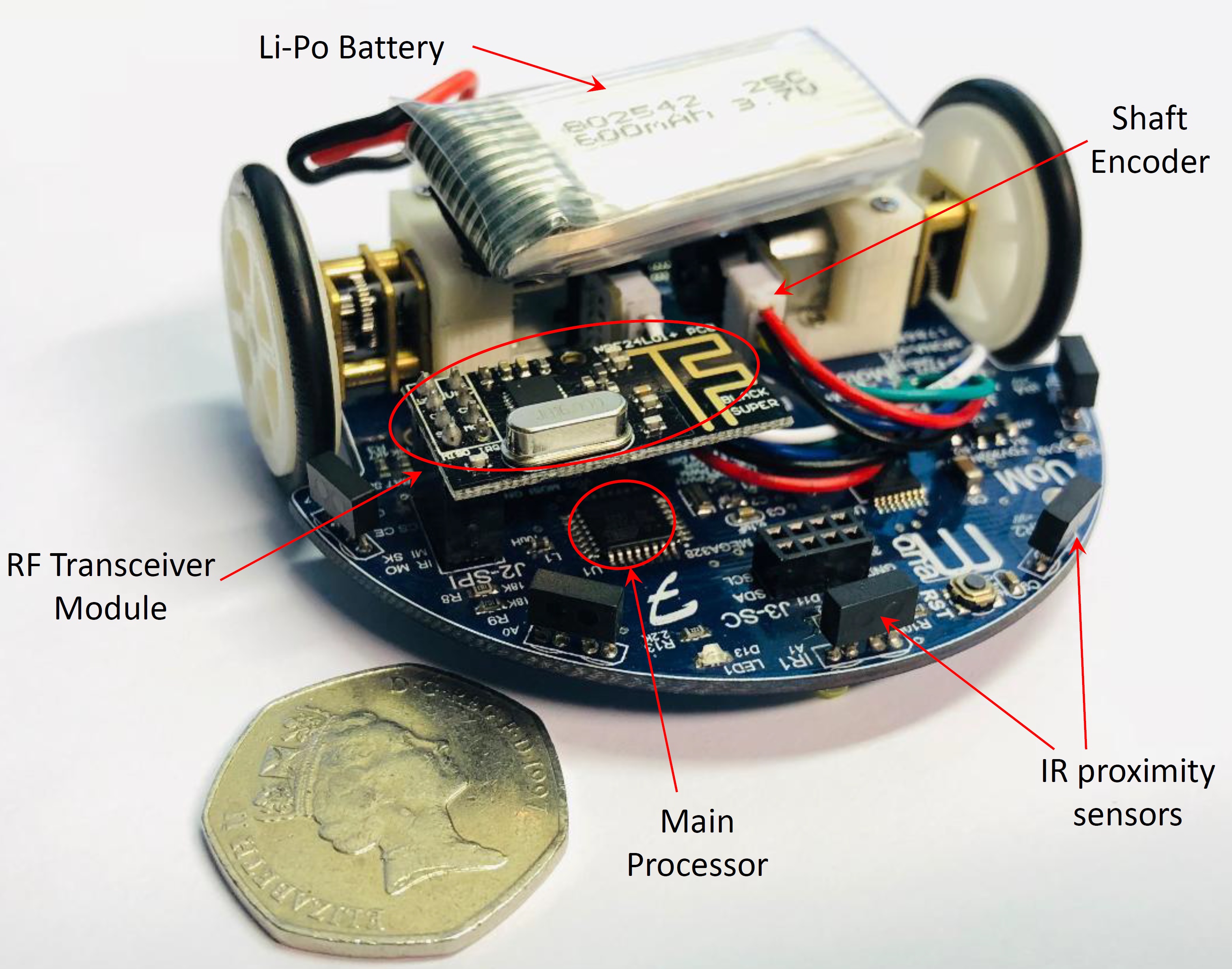}}
\hspace{0.01\linewidth}
\subfloat[]{
\includegraphics[ height=3.5cm, trim={130 130 150 110}, clip]{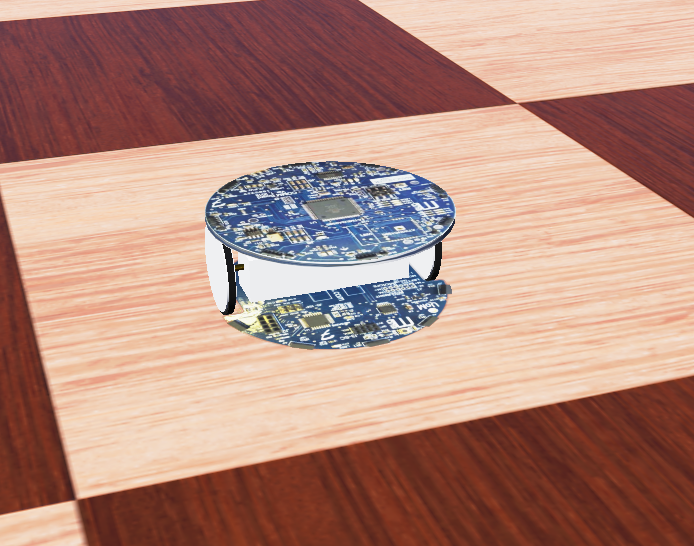}}
\hspace{0.01\linewidth}
\caption{(a) Mona, an open-source miniature mobile robot developed for swarm robotic applications and (b) Mona model in Webots.}
\label{fig:Modeling Mona}
\end{figure}

%
%
To model Mona robot~\cite{arvin2018mona} in Webots, we assembled different parts including upper and lower boards, motors, and wheels which is shown in Fig.~\ref{fig:Modeling Mona}-b. A population of 100 Monas, where presented in the simulations and each one has its own specific ID in order for the \emph{supervisor} to recognize them in the simulation world-model, and calculate the adequate displacement of linear and rotational motion.
Besides the simulations in Webots, we also conducted some simulations in Matlab to check the modifications and to run optimization programs. In this paper, we represented three different setups for simulations in both Webots and Matlab, which will be described in the following subsections.
\subsection{Setup 1: Linear Motion}
\label{subsec:linear motion setup}
The first simulation is executed to prove the linear motion of swarm is achievable thanks to the first modification. The configuration of this setup is similar to Table \ref{table:AES Simulation Settings}, but the rate of rotation is set to zero, $\omega=0$. The result of this Matlab simulation is illustrated in the first row of Fig. \ref{fig:Simulation Results}.
\subsection{Setup 2: Rotational Motion}
\label{subsec:Rotational motion setup}
Similarly, the second modification is verified by another simulation, in which the setting makes the swarm rotate about its center. The setting of Table \ref{table:AES Simulation Settings} is analogous to the current setting, yet the desired linear speed is set to zero, $\lVert{\boldsymbol{\hat{v}}_d=0}\rVert$. The second row of Fig. \ref{fig:Simulation Results} represents a time lapse of this simulation in Matlab. 
\subsection{Setup 3: Combination of Linear \& Rotational Motion}
\label{subsec:linear+Rotational motion setup}
This simulation is conducted in Webots. We considered the robots as independent particles, each of which is determined by its orientation and position in the xy-plane. In this paper, a \emph{supervisor} code is programmed to calculate the position and orientation of robots and dictate them to the corresponding agent. In other words, it is assumed that the controller of robots are ideal and the desired states of robots which supervisor calculates, is the same as the actual state of robots. \par
The settings of this simulation is placed at Table \ref{table:AES Simulation Settings}. By defining such settings, we expect the swarm to rotate about its center while moving collectively toward the west direction. To depict the result of the simulation, we saved a log file from Webots, and then plot the position and rotation of each robots in Matlab, results of which are placed in the third row of Fig. \ref{fig:Simulation Results}.
%
%

\section{Results \& Discussion}
\label{sec:Results}
%

%
Applying TCACS optimization algorithm on the previously defined problem provided us the tuned parameter of the algorithm for a specific setting of simulation. The setting was considered as Table~\ref{table:AES Simulation Settings}. 
In the optimization, the maximum number of function evaluations was limited to 300. Finally, the optimized parameters achieved as $\alpha^*=0.066$, $\beta^*=0.97$, and $k^*=1.28$. The shape of swarm was considered as a square with 10 rows and 10 columns of agents.\par
\begin{table}[!t]
\caption{AES Simulation Settings}
\label{table:AES Simulation Settings}
\centering
  \begin{tabular}{l p{7cm}c } 
    \hline
    \textbf{Parameter$\quad$} 	&\textbf{Description}	& \textbf{Value} 	\\ \hline
    $v_0$ 				& Absolute linear velocity of each agent	& 0.05 [m/sec]		 		\\ 
    $t_f$ 				& Final time of simulation	& 30 [sec]				\\ 
    $N$ 				& Number of agents	& 10*10 			\\ 
    $d_{init}$ 			& Initial distance between two side-by-side agents	& 0.2 [m]			\\ 
    $\boldsymbol{\hat{v}}_d$ 				& Desired linear velocity unit vector	& $[-1.0, 0.0]^T$ [m/sec]	\\ 
    $w_l$ 				& Weight for desired linear velocity	& 0.8					\\ 
    $\omega$ 			& Desired rate of rotation	& 0.7 [rad/sec]					\\ 
     $w_r$ 				& Weight for desired rotational velocity	& 1					\\ 
    $D_r$ 				& Strength coefficient of measurement noise	& 0.5 				\\ 
    $D_{\theta}$ 		& Strength coefficient of actuation noise	& 0.02 				\\ 
    \hline
  \end{tabular}
\end{table}
%
%
%
%
%
%
\newcommand\figOptWidth{1.7}
\begin{figure}%
    \centering
    \subfloat[]{{\includegraphics[width=\figOptWidth in]{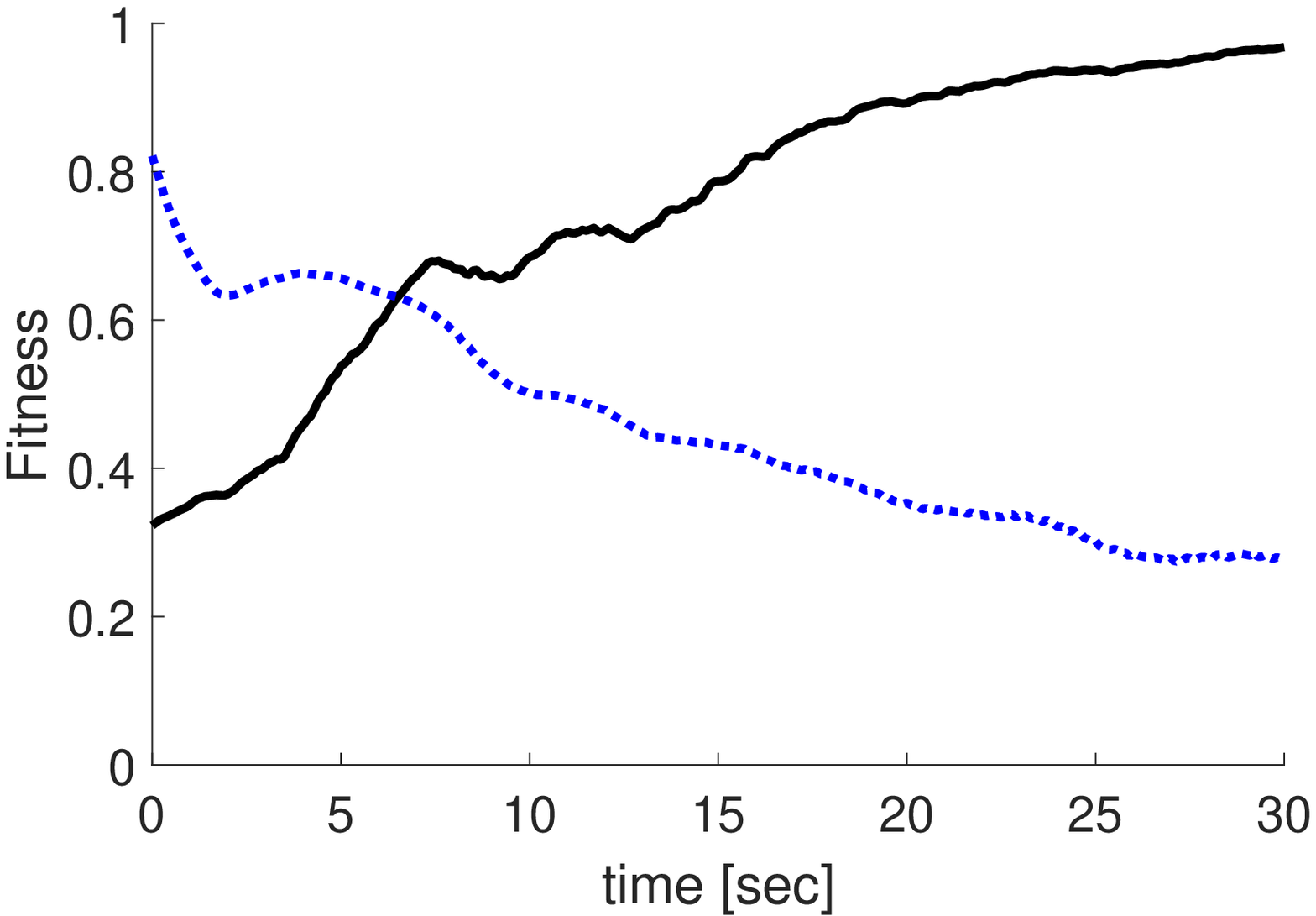} }} 
    \subfloat[]{{\includegraphics[width=\figOptWidth in]{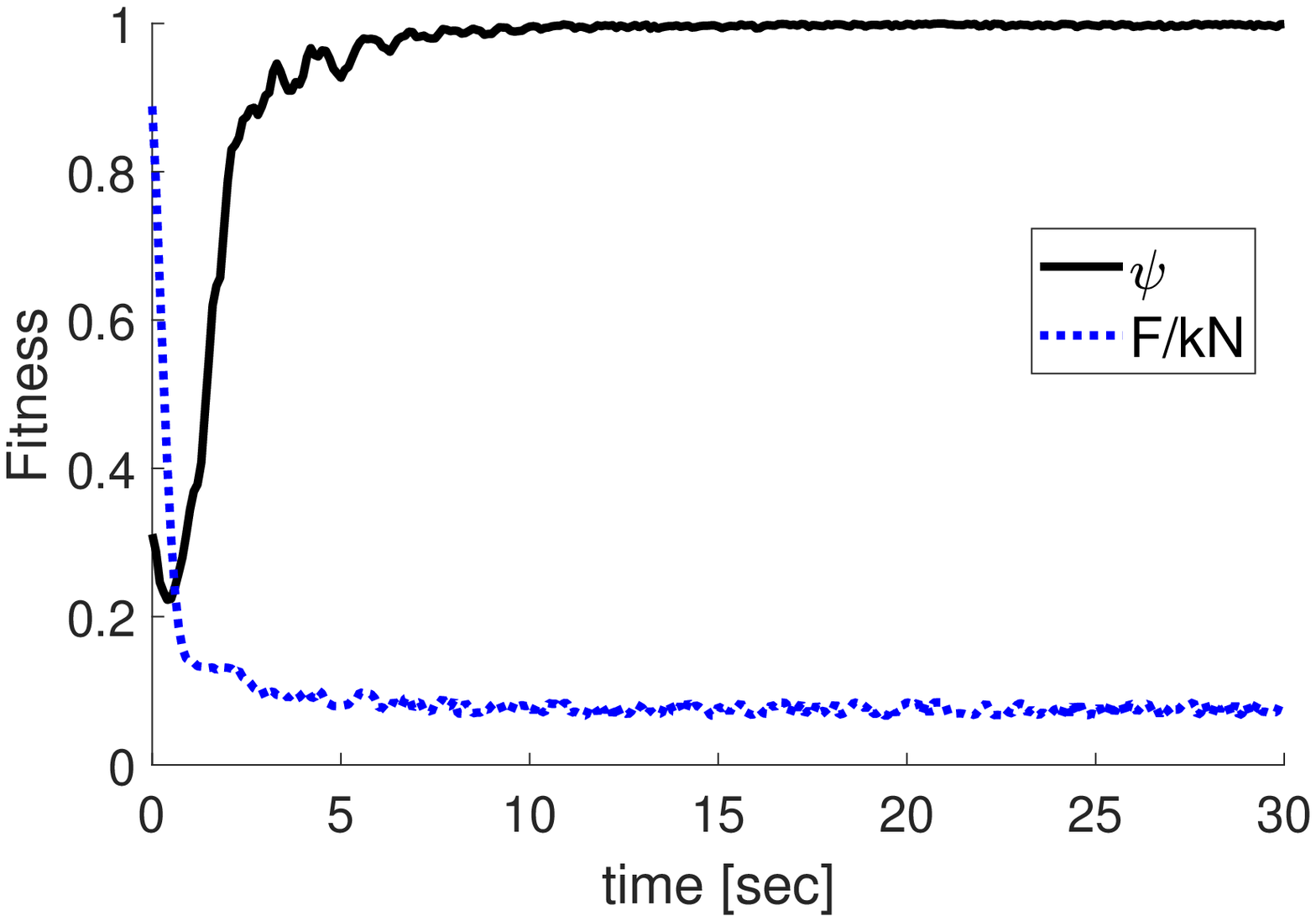} }} 
    
    \subfloat[]{{\includegraphics[width=\figOptWidth in]{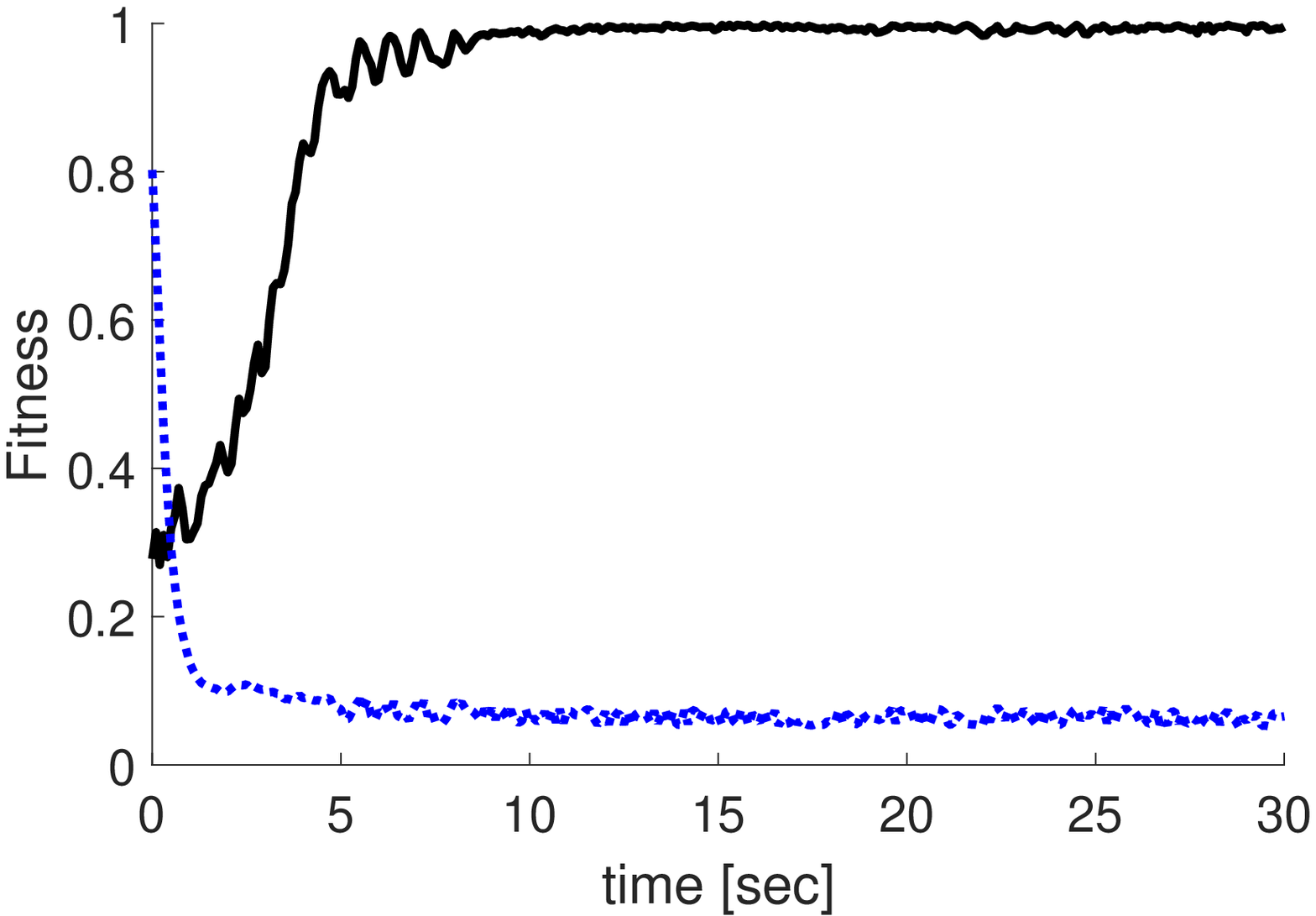} }} 
    \subfloat[]{{\includegraphics[width=\figOptWidth in]{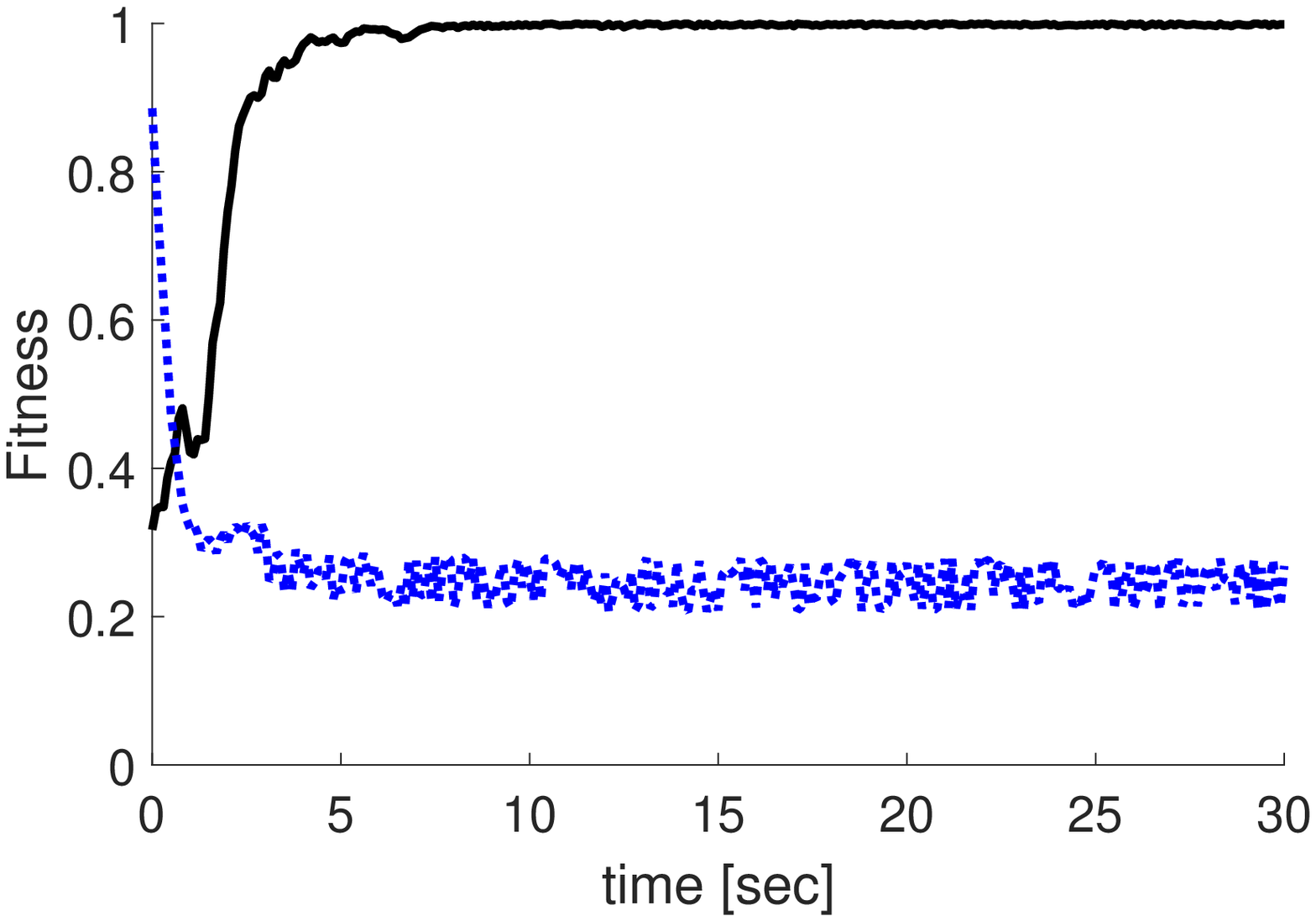} }} 
    
    \caption{Effect of Parameter optimization, and Weights on CM: a) Default parameters, b) optimized parameters, c) optimized parameters with emphasis on force objective, d) optimized parameters with emphasis on alignment objective}%
    \label{fig:Effect of Opt}%
\end{figure}
In order to show the effect of optimized parameters on the performance of flocking, we investigated those objectives with both the optimized and non-optimized set of parameters. Fig.~\ref{fig:Effect of Opt}-b shows the significant improvements, in comparison with the default parameters Fig.~\ref{fig:Effect of Opt}-a, which is achieved by optimized parameters. As it is shown, the rate of convergence is increased, so the CM achieved faster. Besides, the total force of the network is decreased, meaning that the shape of the swarm is maintained better. \par
We also studied the effect of objective weights $w_1$, and $w_2$ on the optimized parameters. Since the behavior of swarm varies for each obtained parameters, we defined two more different weight sets to show the effect of different weights on swarm behavior: 
\begin{itemize}
    \item More emphasis on the first objective, $w_1=10/N$, $w_2=-1$ (Fig.~\ref{fig:Effect of Opt}-c)
    \item More emphasis on the second objective, $w_1=1/N$, $w_2=-10$ (Fig.~\ref{fig:Effect of Opt}-d)
\end{itemize}
As we expected, the former setting resulted in the more reduction of total force, and the latter one, increased the rate of convergence of alignment, yet made the system more sensitive to noise (the effect of which is shown on the fluctuating behavior of force on Fig.~\ref{fig:Effect of Opt}-d). 
%
%
%
%
%
The first row of Fig. \ref{fig:Simulation Results} proves that the initially perturbed formation of the swarm is eliminated, then the collective motion of swarm is achieved. The inducing linear force caused the swarm to move toward the west direction, as it was expected by $\boldsymbol{\hat{v}}_d$. Due to the robustness of AES to noises, noises were not effective enough to prevent the swarm from flocking. The time history of the second row, for rotational motion, proves that the second modification is verified, similarly. In this simulation, the swarm rotates about its center and maintain its shape, meanwhile.\par
The third row shows that the combination of linear and rotational motions made the swarm to move toward the desired direction, and in the mean time, rotate about its center. The same optimized parameters were used for this simulation, in Webots, proving that similar settings results in similar behavior in Matlab and Webots.
%
%
\newcommand\figWidth{1.1}
\begin{figure}%
    \centering
    \subfloat{{\includegraphics[width=\figWidth in]{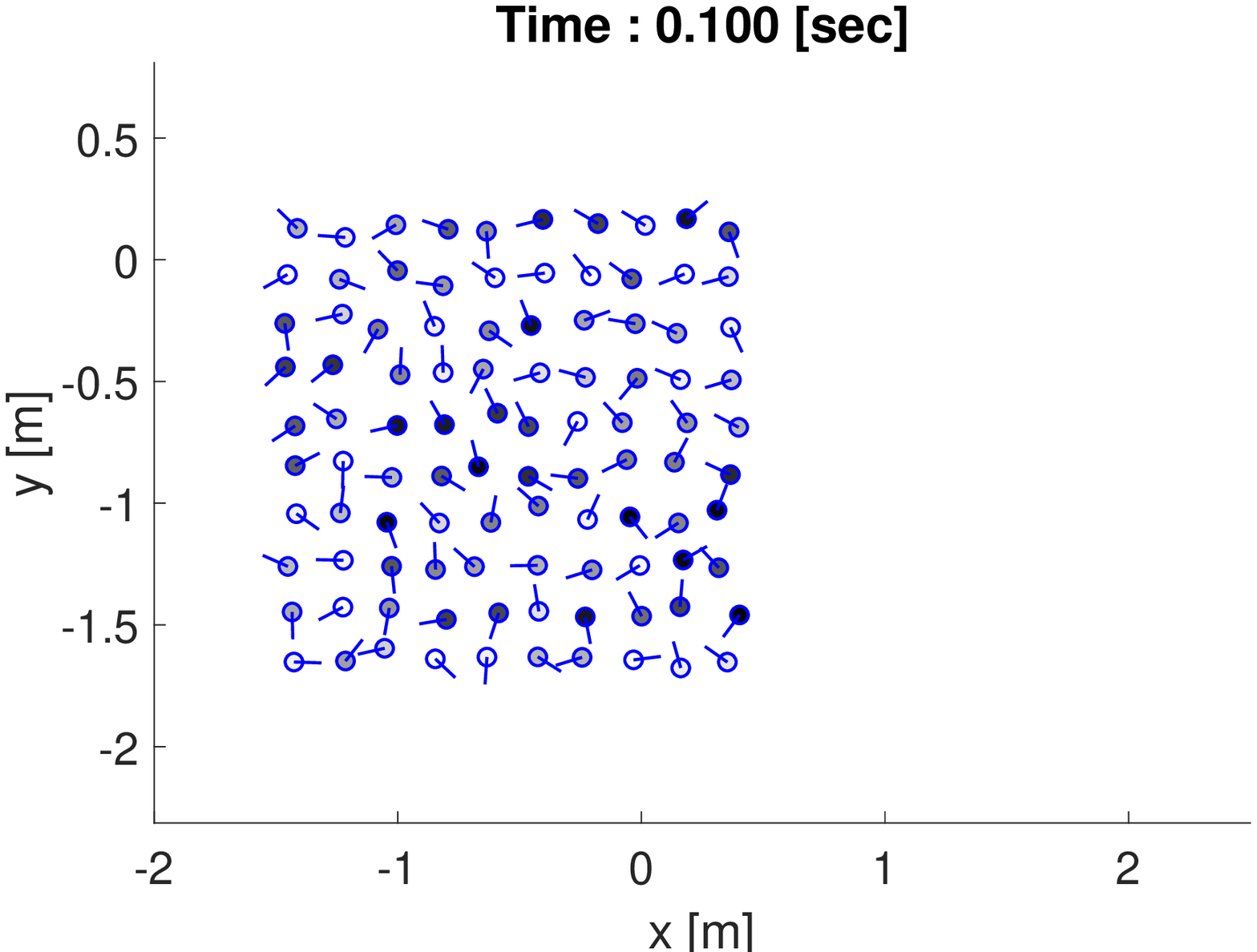} }} 
    \subfloat{{\includegraphics[width=\figWidth in]{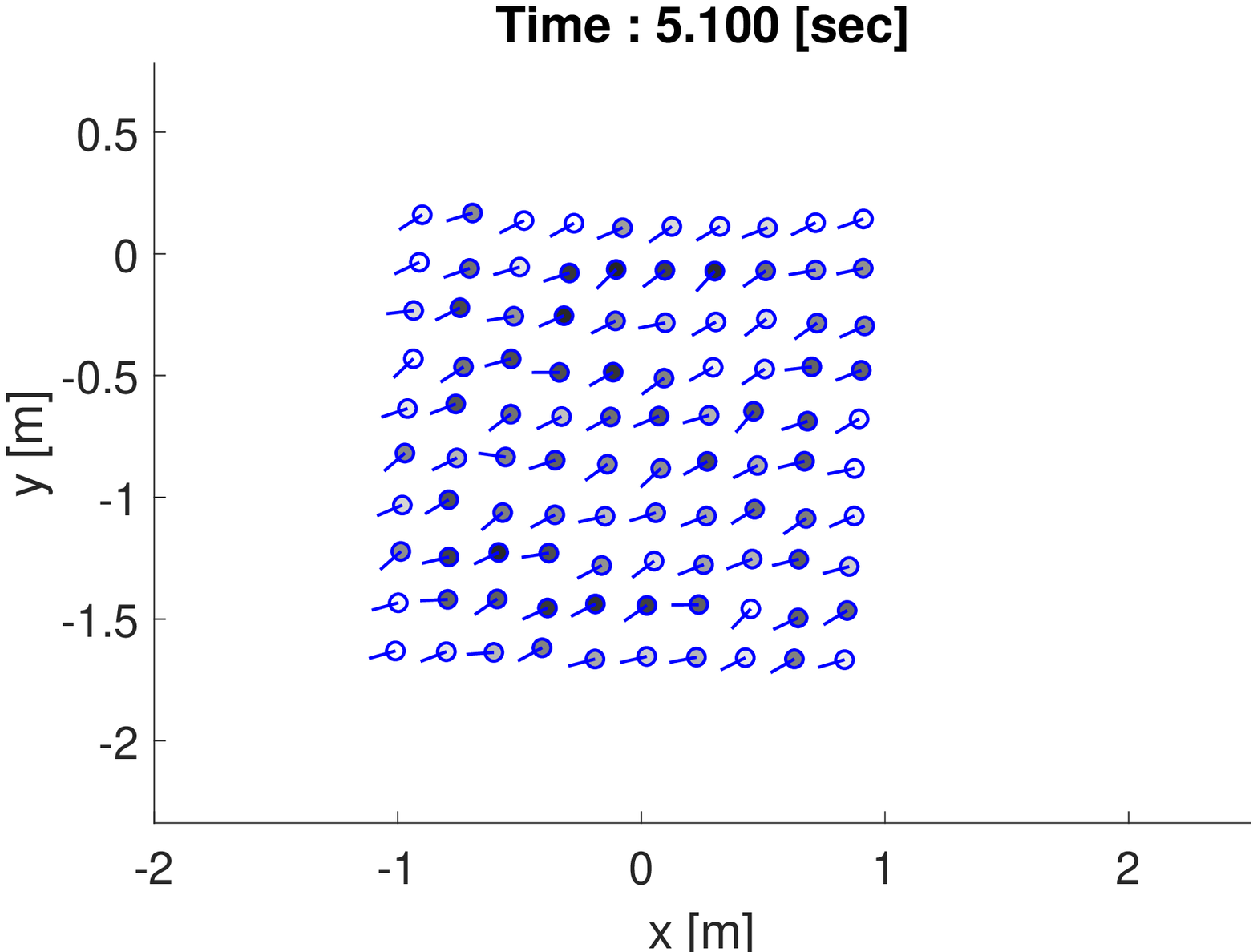} }} 
    \subfloat{{\includegraphics[width=\figWidth in]{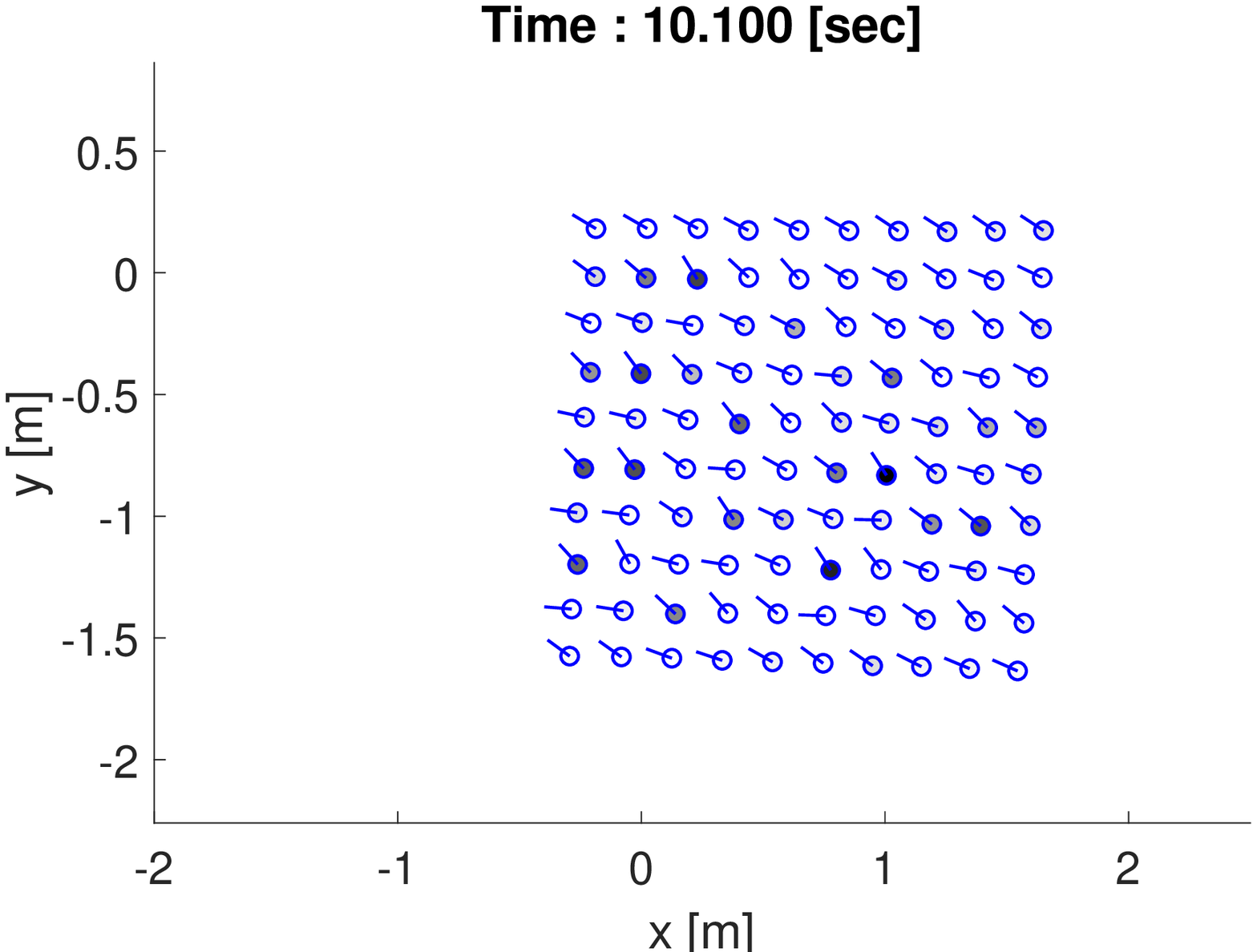} }} 
    \subfloat{{\includegraphics[width=\figWidth in]{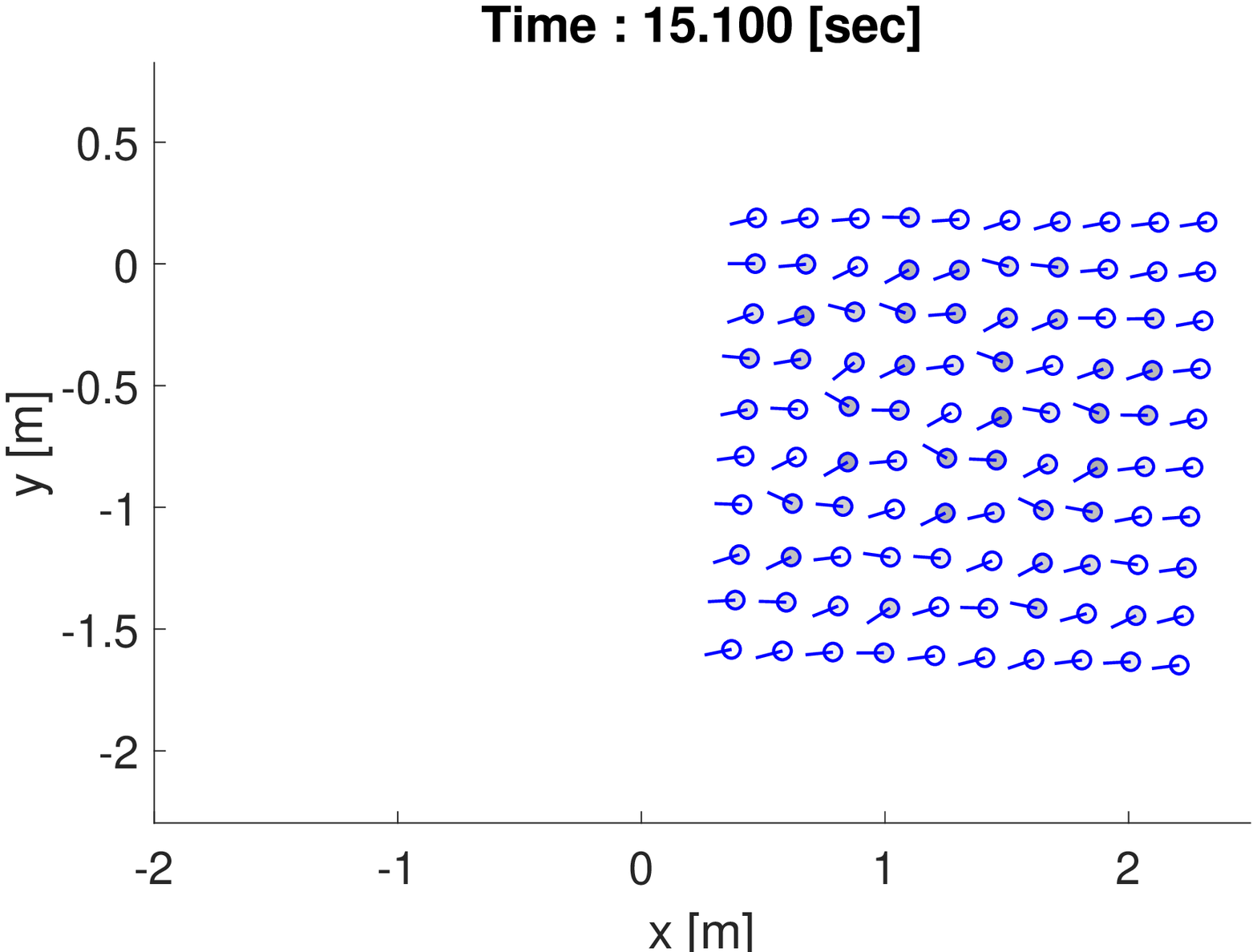} }} 

    \subfloat{{\includegraphics[width=\figWidth in]{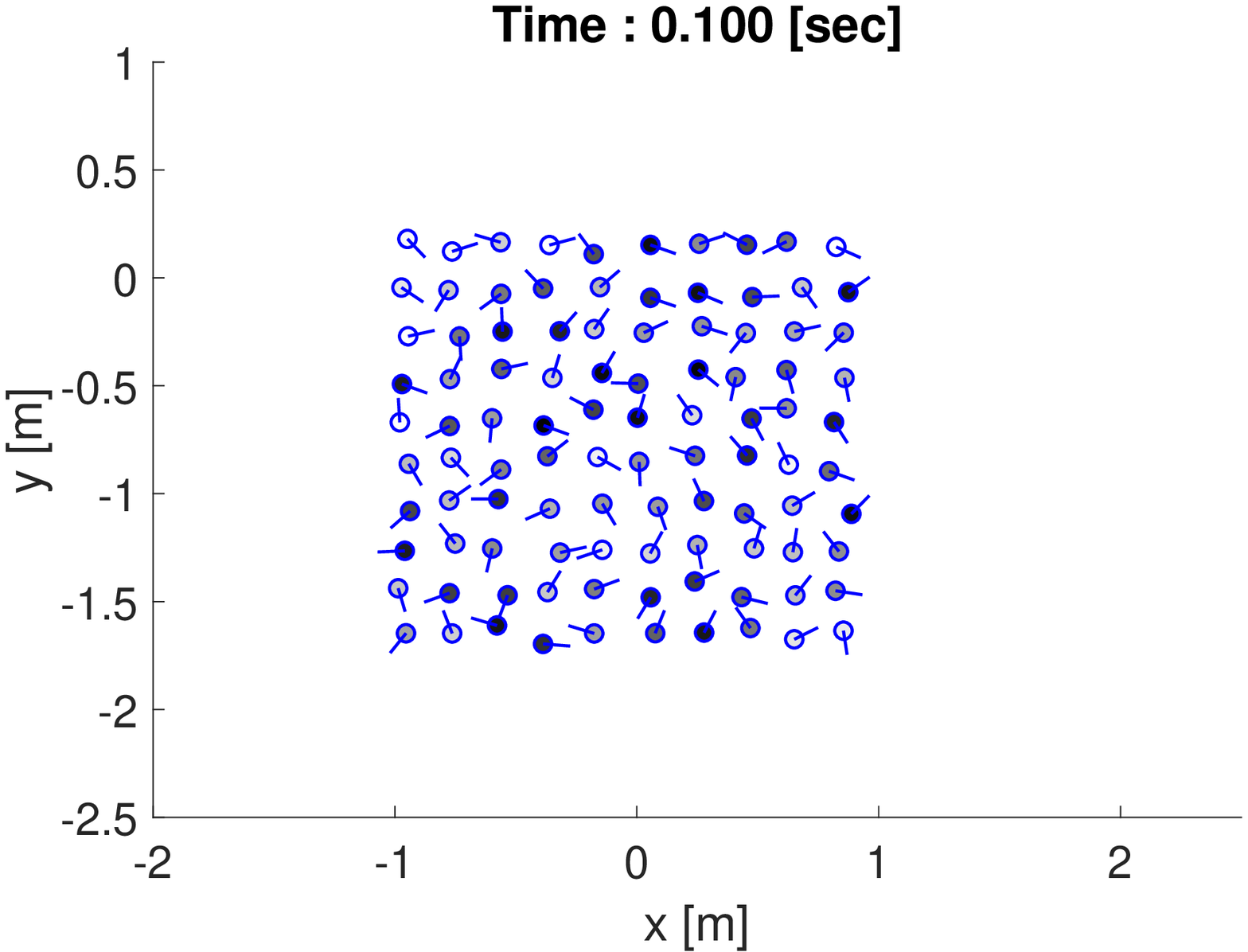} }} 
    \subfloat{{\includegraphics[width=\figWidth in]{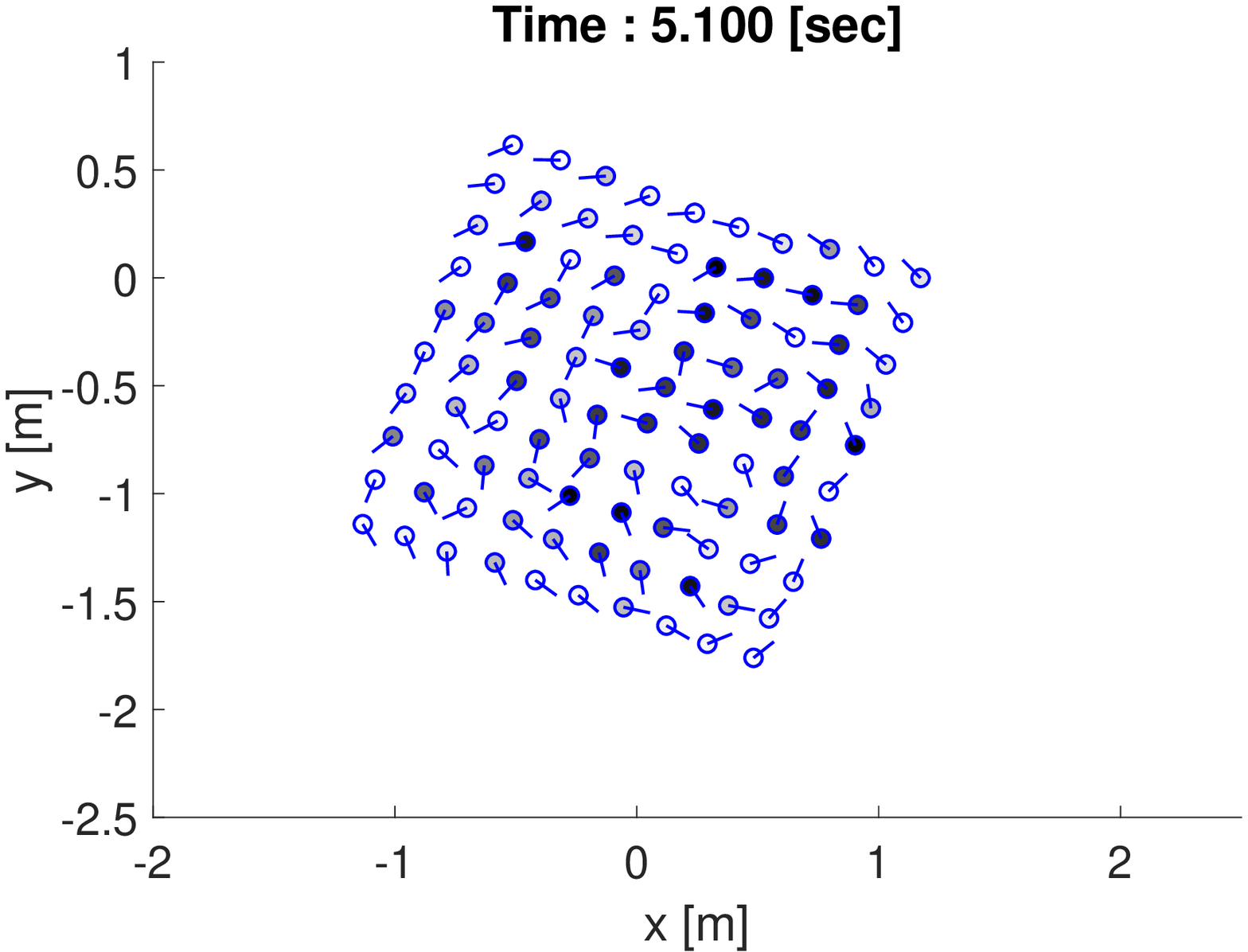} }} 
    \subfloat{{\includegraphics[width=\figWidth in]{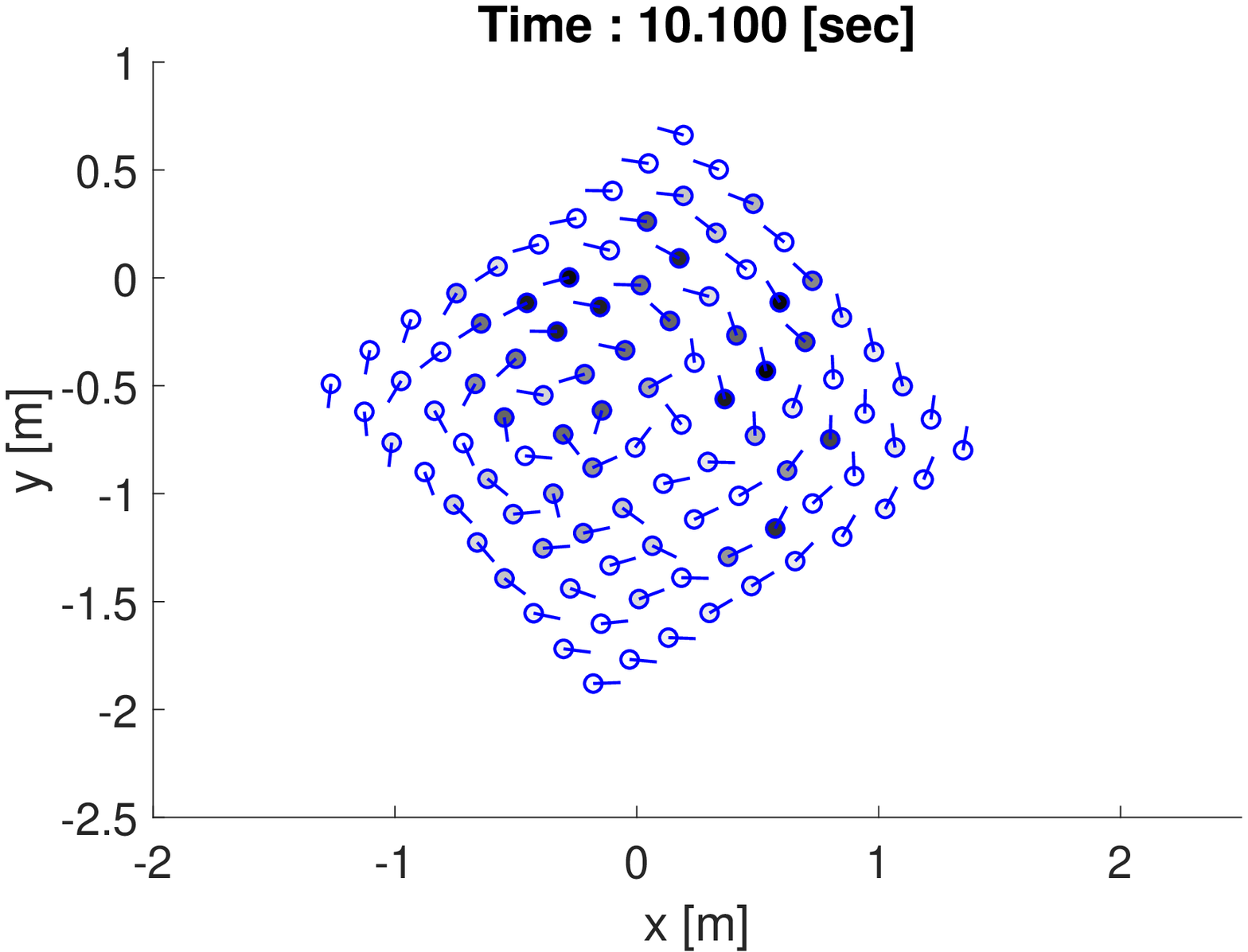} }} 
    \subfloat{{\includegraphics[width=\figWidth in]{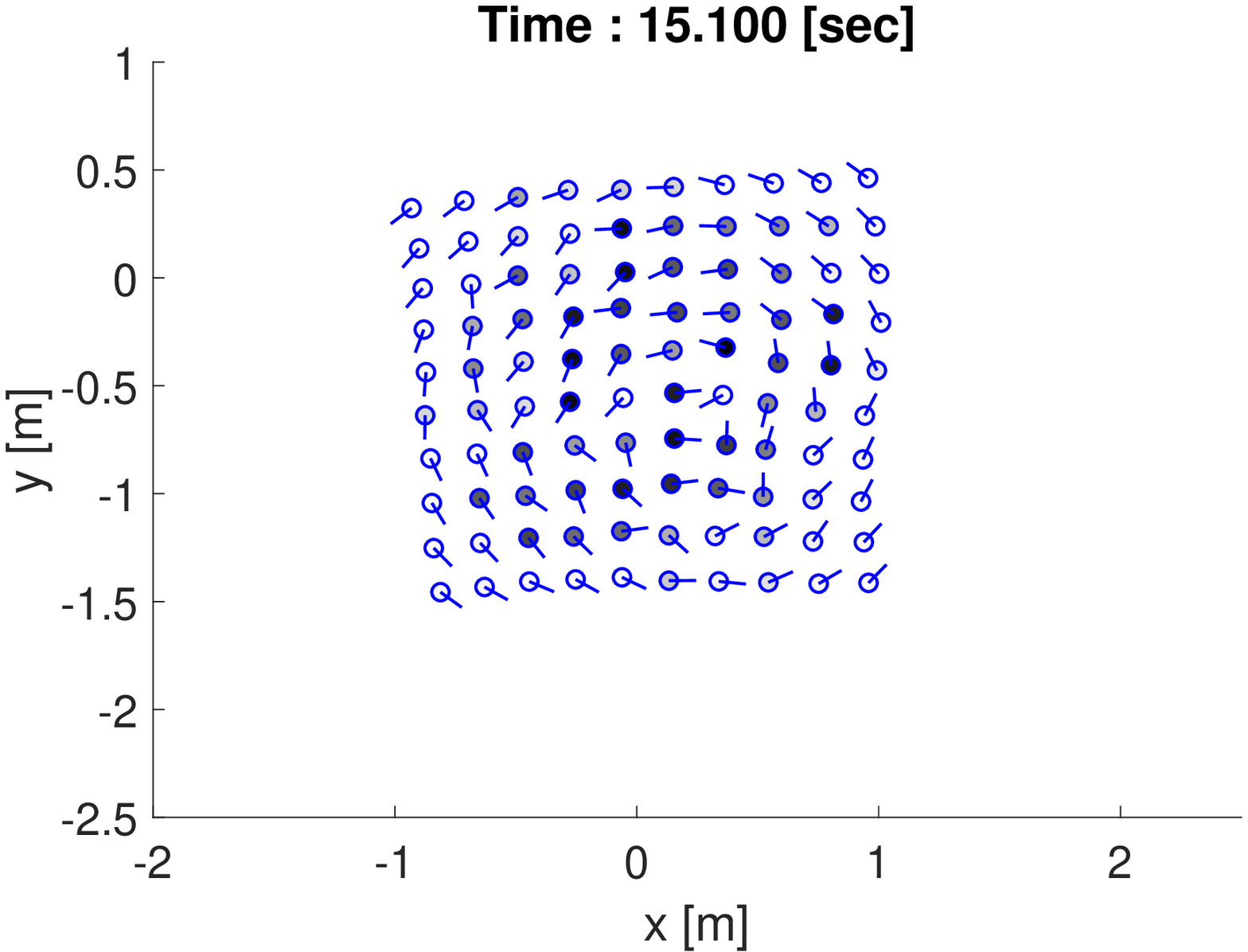} }} 

    \subfloat{{\includegraphics[width=\figWidth in]{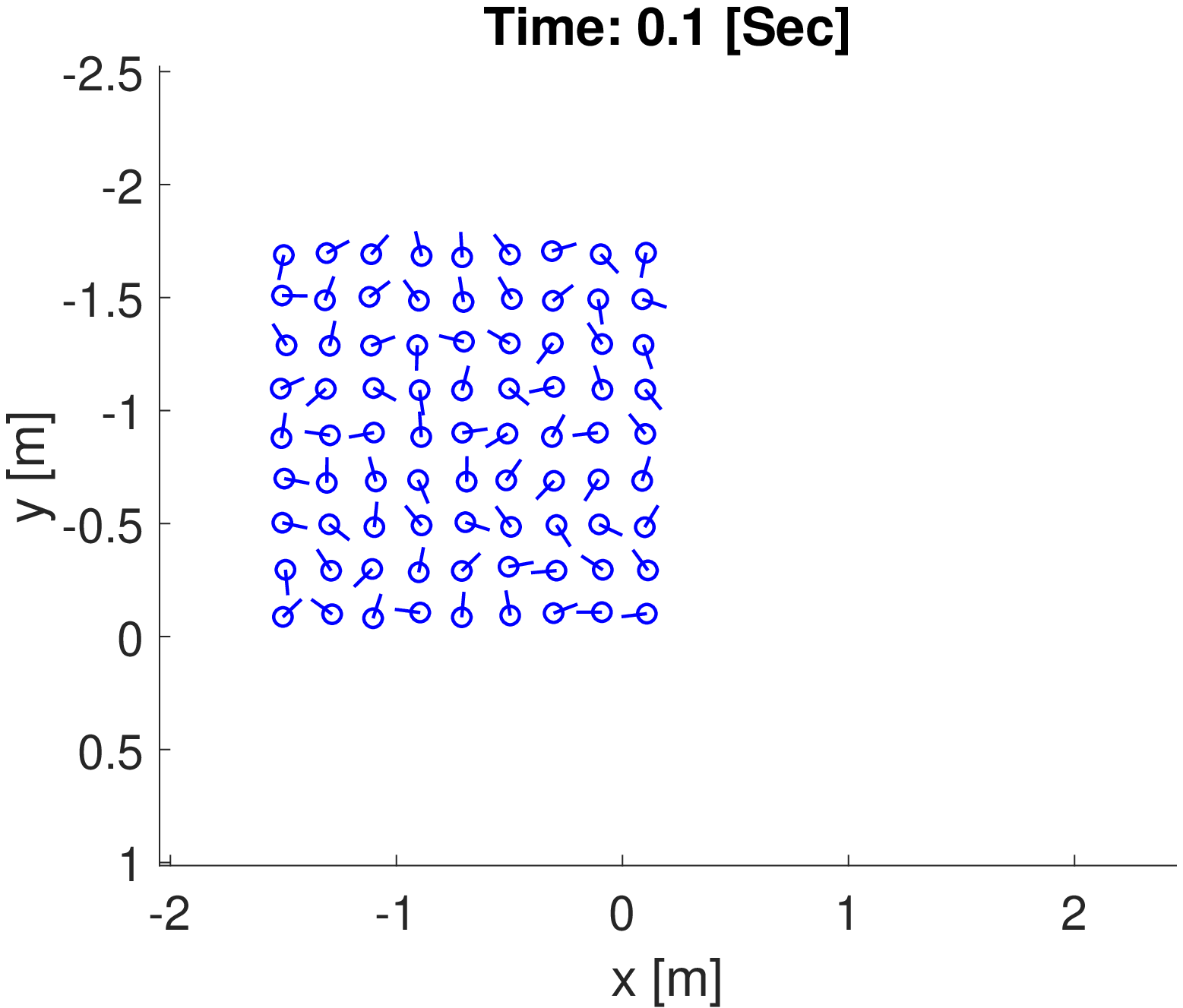} }} 
    \subfloat{{\includegraphics[width=\figWidth in]{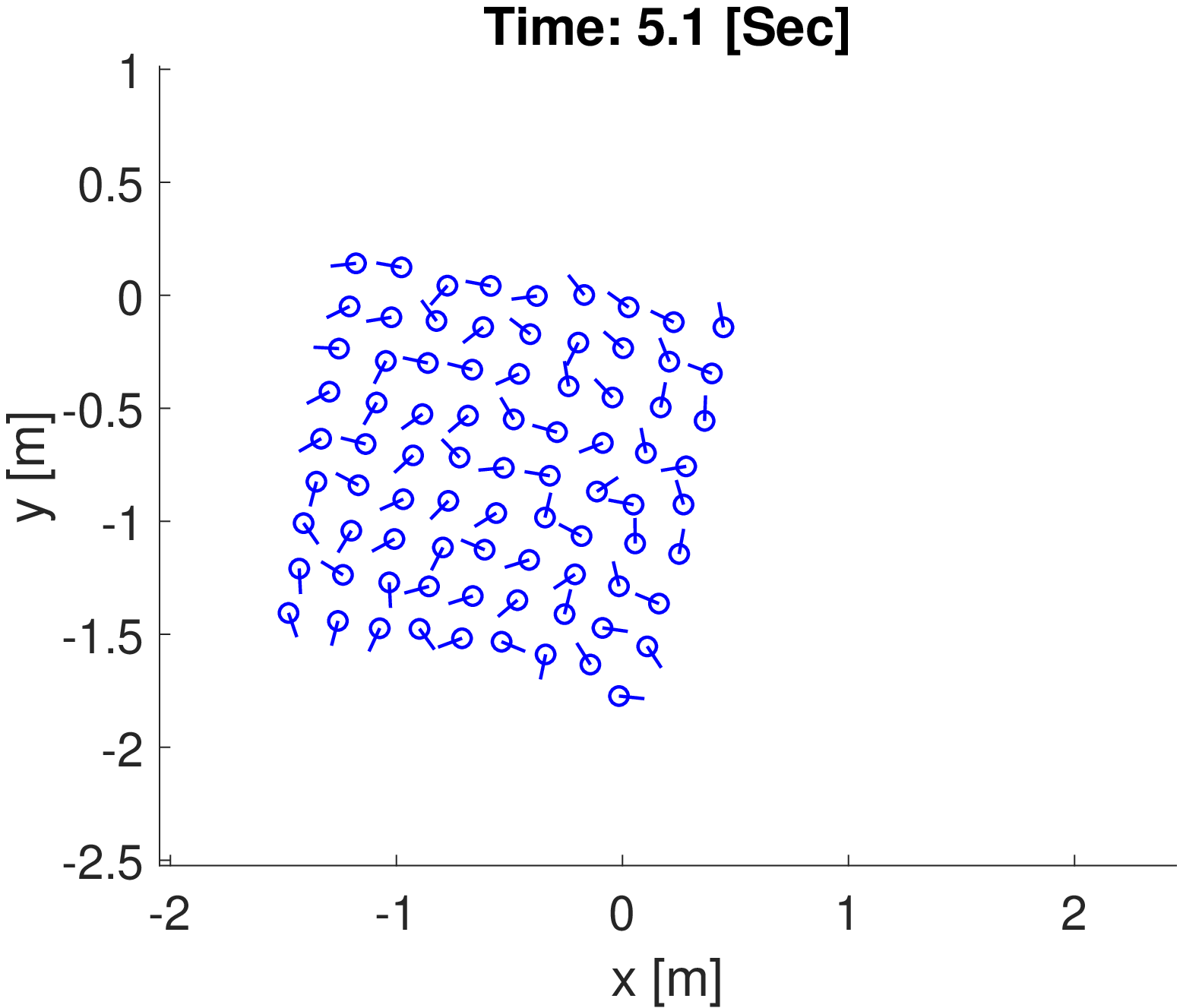} }} 
    \subfloat{{\includegraphics[width=\figWidth in]{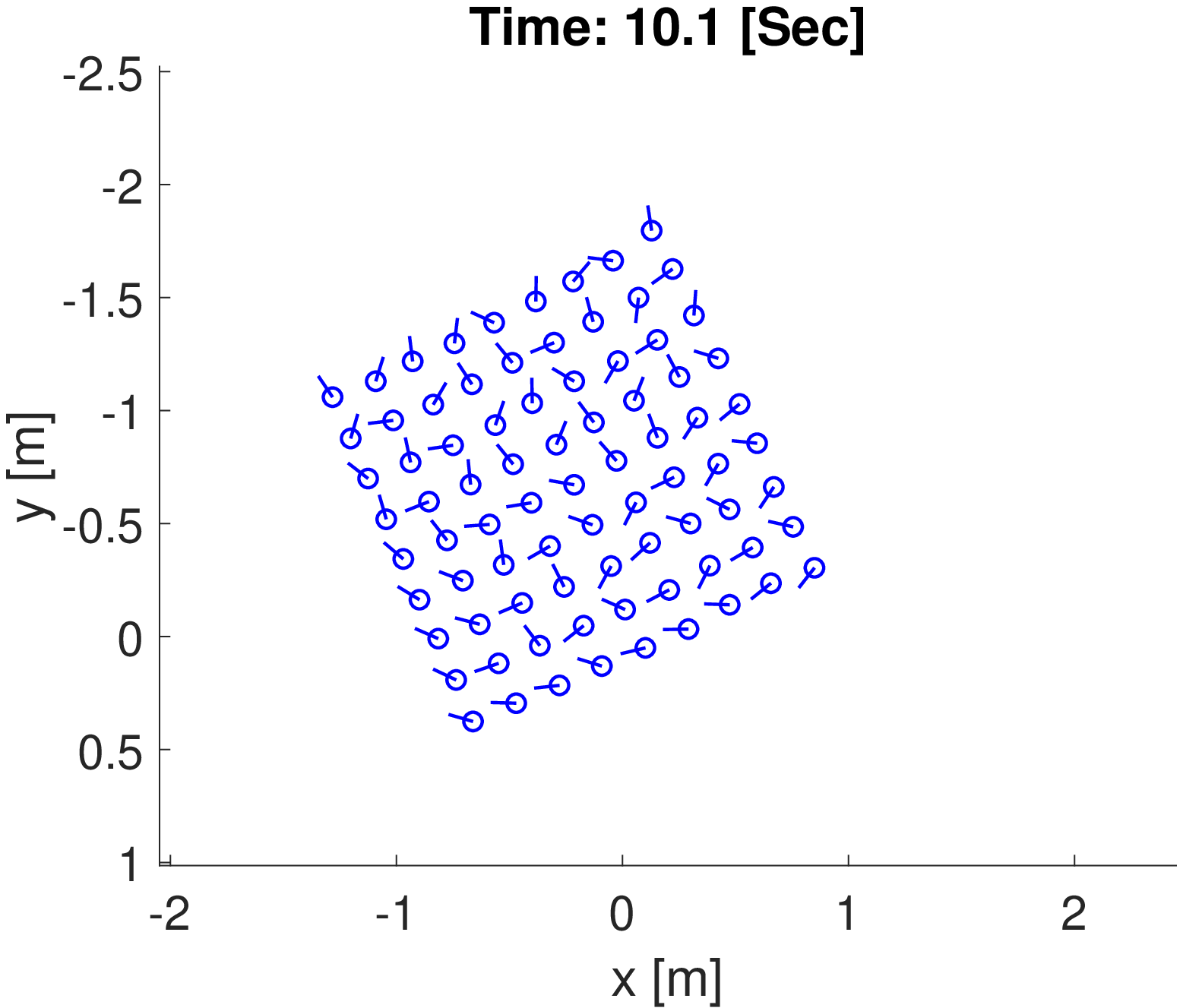} }} 
    \subfloat{{\includegraphics[width=\figWidth in]{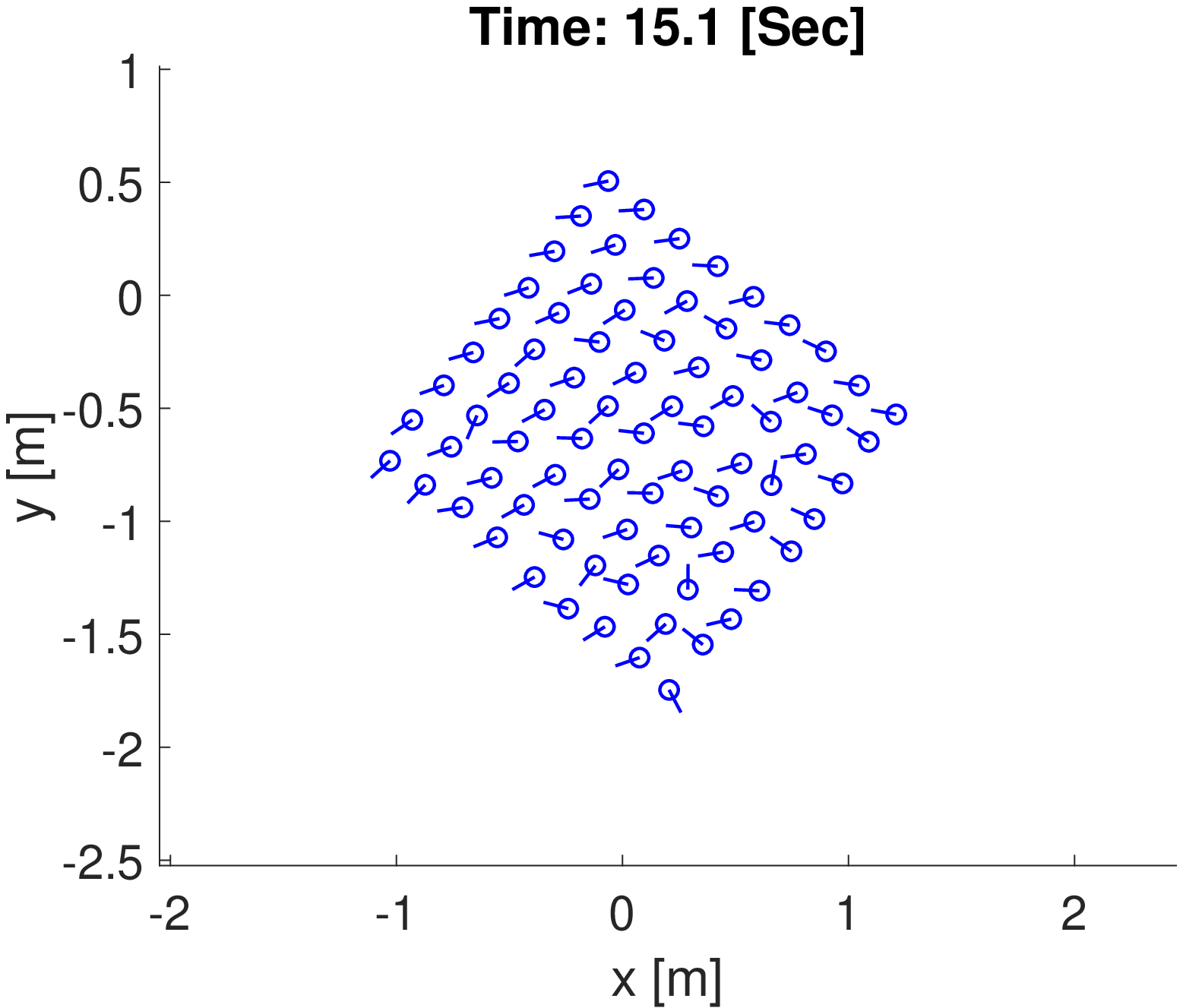} }} 
    
    
    \caption{Simulation Results: First row: Pure linear motion in Matlab, Second row: Pure rotational motion in Matlab, Third row: Linear+rotational motion in Webots}%
    \label{fig:Simulation Results}%
\end{figure}
\section{Conclusion}
\label{sec:Conclusion}
We have addressed the parameter optimization and Webots simulations of AES model, which is an elasticity-based approach for achieving CM. Some of its advantages such as not relying on exchanging orientation information, and assuming 2-DoF agents attracted our attention. So, considering our robotic case, we added some modifications helping the swarm of robots to move toward the desired direction and turn about an arbitrary rotation center point. In addition, a new criterion for measuring the alignment of the swarm is defined based on the desired direction of motion. 
\par
utilizing TCACS optimization algorithm, we tuned the parameter of AES model for a specific setup containing combination of linear and rotational motion of swarm. The effect of optimization on the behavior of CM, such as increasing the rate of convergence and decreasing the total force, has been shown. Various weighting strategies for optimization has been studied, too. \par
Finally, three different scenarios have been conducted in Matlab and Webots. For this end, we have designed a CAD model of Mona robot in Webots and have simulated a supervised program to control agents. The results proved that our modifications are applicable and the optimized parameters improved the performance of the AES mechanism.
%
%
%
%
%
%
%
\bibliographystyle{splncs04}

 \bibliography{mybibliography.bib}

\end{document}